\documentclass[10pt,twocolumn,letterpaper]{article}

\usepackage{iccv}
\usepackage{times}
\usepackage{epsfig}
\usepackage{graphicx}
\usepackage{amsmath}
\usepackage{amssymb}
\usepackage{paralist}
\usepackage{xcolor}
\usepackage{color, colortbl}

\usepackage[pagebackref=true,breaklinks=true,letterpaper=true,colorlinks,bookmarks=false]{hyperref}

\iccvfinalcopy % *** Uncomment this line for the final submission

\def\iccvPaperID{7034} % *** Enter the ICCV Paper ID here

\ificcvfinal\fi

\definecolor{LightCyan}{rgb}{0.88,1,1}

\newcommand{\mypartitle}[2][2.5]{\vspace*{-#1 ex}~\\{\noindent {\bf #2}}}

\newcommand{\inputimage}{\ensuremath{\mathbf{I}}\xspace}
\newcommand{\inputheadbbox}{\ensuremath{\mathbf{h}_{\text{bbox}}}\xspace}
\newcommand{\inputheadcrop}{\ensuremath{\mathbf{h}_{\text{crop}}}\xspace}

\newcommand{\Npersons}{\ensuremath{N_p}\xspace}
\newcommand{\Nimagetokens}{\ensuremath{N}\xspace}
\newcommand{\Nalltokens}{\ensuremath{N_t}\xspace}

\newcommand{\seqimgpatches}{\ensuremath{\mathbf{s}_{p}^{\text{img}}}\xspace}
\newcommand{\seqimgtokens}{\ensuremath{\mathbf{s}_{t}^{\text{img}}}\xspace}
\newcommand{\gazetoken}{\ensuremath{\mathbf{x}^{\text{emb}}}\xspace}
\newcommand{\imgtokens}{\ensuremath{\mathbf{x}^{\text{img}}}\xspace}
\newcommand{\persontoken}{\ensuremath{\mathbf{x}^{\text{g}}}\xspace}
\newcommand{\globaltoken}{\ensuremath{\mathbf{x}^{\text{glo}}}\xspace}

\newcommand{\token}{\ensuremath{\mathbf{x}\xspace}}

\newcommand{\inputtokens}{\ensuremath{\mathbf{x}}\xspace}
\newcommand{\outputtokens}{\ensuremath{\mathbf{x}^{(L)}}\xspace}
\newcommand{\outtokens}{\ensuremath{\mathbf{x}^{\text{out}}\xspace}}

\newcommand{\gazepoint}{\ensuremath{\mathbf{g}_{\text{pt}}}\xspace}
\newcommand{\gazevector}{\ensuremath{\mathbf{g}_{\text{v}}}\xspace}

\newcommand{\imageproj}{\ensuremath{\mathcal{P}_{\text{img}}}\xspace}
\newcommand{\bboxproj}{\ensuremath{\mathcal{P}_{\text{bbox}}}\xspace}
\newcommand{\gazeproj}{\ensuremath{\mathcal{P}_{\text{gaze}}}\xspace}
\newcommand{\gazepred}{\ensuremath{\mathcal{P}_{\text{gpred}}}\xspace}
\newcommand{\mlpdecoder}{\ensuremath{\mathcal{D}_{\text{MLP}}}\xspace}
\newcommand{\gazebackbone}{\ensuremath{\mathcal{G}}\xspace}

\newcommand{\bboxemb}{\ensuremath{\mathbf{x}^{\text{bbox}}}\xspace}
\newcommand{\gazeemb}{\ensuremath{\mathbf{g}^{\text{emb}}}\xspace}

\newcommand{\inoutlabel}{\ensuremath{\mathbf{o}}\xspace}
\newcommand{\heatmap}{\ensuremath{\mathcal{A}}\xspace}

\newcommand{\inoutmlp}{\ensuremath{\mathcal{O}_\text{MLP}}\xspace}

\newcommand{\loss}{\ensuremath{\mathcal{L}}\xspace}
\newcommand{\losscoeff}{\ensuremath{\mathcal{\lambda}}\xspace}

\begin{document}

%%%%%%%%% TITLE
\title{Sharingan: A Transformer-based Architecture for Gaze Following}

\author{Samy Tafasca \and Anshul Gupta \and Jean-Marc Odobez
\and Idiap Research Institute, Switzerland\\
École Polytechnique Fédérale de Lausanne, Switzerland\\
{\tt\small \{stafasca, agupta, odobez\}@idiap.ch}
% For a paper whose authors are all at the same institution,
% omit the following lines up until the closing ``}''.
% Additional authors and addresses can be added with ``\and'',
% just like the second author.
% To save space, use either the email address or home page, not both
}

\maketitle
% Remove page # from the first page of camera-ready.
\ificcvfinal\thispagestyle{empty}\fi

%%%%%%%%% ABSTRACT
\begin{abstract}
    \label{sec:abstract}

%Abstract
Gaze is a powerful form of non-verbal communication and social interaction that humans develop from an early age. As such, modeling this behavior is an important task that can benefit a broad set of application domains ranging from robotics to sociology. In particular, Gaze Following is defined as the prediction of the pixel-wise 2D location where a person in the image is looking. Prior efforts in this direction have focused primarily on CNN-based architectures to perform the task. In this paper, we introduce a novel transformer-based architecture for 2D gaze prediction. We experiment with 2 variants: the first one retains the same task formulation of predicting a gaze heatmap for one person at a time, while the second one casts the problem as a 2D point regression and allows us to perform multi-person gaze prediction with a single forward pass. This new architecture achieves state-of-the-art results on the GazeFollow and VideoAttentionTarget datasets. The code for this paper will be made publicly available.
\end{abstract}

%%%%%%%%% INTRODUCTION
\section{Introduction}
\label{sec:introduction}

% Introduction

Gaze is an important form of human communication and was extensively studied across different domains and applications such as consumer behavior understanding \cite{behe2020seeing, tomas2021goo, kim2021assessing}, sociology by analyzing different gaze behaviors (\eg joint attention, eye contact) \cite{fan2018inferring_videocoatt, marin2014detecting, marin2019laeo}, robotics through human-robot interactions \cite{sheikhi2015combining, jin2022depth, admoni2017social} and clinical research for the study of neurodevelopmental disorders \cite{chong2020dvisualtargetattention, li2022appearance} to cite a few.

Unlike traditional works on gaze analytics proposed by the computer vision community which focused mainly on predicting gaze directions (\ie 3D angular values) from the eyes \cite{yu2018deep} or the face \cite{kellnhofer2019gaze360} of a person, gaze following \cite{nips15_recasens} tackles the task in a more general form where the goal is to infer the 2D location in the image where a person is looking without the need for any assumptions or wearable devices. This formulation is particularly interesting in the context of analyzing social scenes and human interactions given the important role that gaze behavior plays in social dynamics. 

In this work, we are mainly interested in addressing the gaze following task using a novel and flexible architecture that can later be extended to incorporate more information in order to analyze scenes featuring human interactions. 
Tasks such as Human-Human-Object Interaction detection \cite{orcesi2021detecting} are particularly relevant for this end goal. Graph neural networks \cite{wu2019comprehensive} achieved 
great success in this area \cite{liang2021visual, qi2018learning} by representing the scene as a graph where nodes denote people or objects, and edges denote the relationship between them. 
This allows the direct exchange of joint interactive information between nodes, irrespective of the distance between them. 

The idea of using a graph-based method to infer possible interactions between people and objects was also proposed for gaze prediction \cite{hu2022gaze}.
However, while graph neural networks are largely flexible, the main shortcoming with their formulation in interaction understanding is the need for strong off-the-shelf object detectors that can accurately and reliably identify the different people and relevant objects in a given image. 
Moreover, object detectors typically do not include non-countable objects (\eg wall, floor, ocean, road), although one might still be interested in identifying the 2D gaze target location within such regions (\eg a position on a white board).
%which can also be valid gaze targets. 
This is arguably the reason why the authors of \cite{hu2022gaze} decided to retain the entire scene image as input for the gaze prediction stage, in order to fill in the missing blanks, 
restraining the use of the graph neural network as a mean 
to compute an interaction heatmap passed along with the image and highlighting the different objects a target person might be gazing at. 

In order to account for all relevant entities in the scene, we turn our attention to transformers \cite{vaswani2017attention} which are graph-like architectures where all tokens interact with each other through an attention mechanism. 
This setup allows us to define a novel \emph{person gaze token} to represent a given person in the scene, which is simply added to the set of image tokens. 
This approach can be simply extended to include as many gaze tokens 
as there are people in the scene: 
this allows our approach to not only model how the gaze information 
of one person interacts with the scene to identify salient gaze targets
for that person, but also to consider and mode the possible interactions between them, like looking at each other or shared-attention, 
and to make it easy to predict their gaze targets in a single forward pass.

While the initial formulation of the person gaze token in this paper only encodes gaze and head location information, it can easily be extended in future works to integrate other multimodal cues, and possibly predict multiple outputs paving the way to a foundational model for social scene understanding. 

\noindent The contributions of this paper are summarized below:
\begin{itemize}
    \item We propose and motivate a novel transformer-based architecture for the gaze following task and achieve state-of-the-art results on available public benchmarks;
    \item We introduce two variants: the first one retains the traditional heatmap prediction task formulation, while the other casts the problem as a 2D point regression;
    \item We show that the second variant is able to perform multi-person gaze prediction in an accurate and effective way;
    \item We also find that this variant of the architecture benefits, performance-wise, from the interaction that follows from processing multiple people at the same time.
\end{itemize}
Experiments on two public benchmark datasets demonstrate the validity of our approach.

%%%%%%%%% RELATED WORK
\section{Related Work}
\label{sec:related-work}

% Related Work
In this section, we present several research areas related to our Sharingan architecture.

\textbf{Gaze Following.}
The task of gaze following was first introduced in the seminal work of Recasens \emph{et al.} \cite{nips15_recasens}. The idea is to predict the pixel-wise 2D location in the image corresponding to where a target person is looking within the scene. The main advantage of this formulation is the lack of constraints which allows methods trained this way to generalize to arbitrary settings (\ie scene properties, camera parameters, image conditions, etc.). It was later extended by Chong \emph{et al.} \cite{chong2020dvisualtargetattention} to also include the prediction of whether the given person is looking inside the image frame or somewhere outside. 

Traditional methods for gaze following \cite{nips15_recasens, chong2020dvisualtargetattention, Fang_2021_CVPR_DAM, gupta2022modular, jin2021multi, lian2018believe, jin2022depth} typically rely on convolutional networks and follow a 2-tower architecture. The first branch processes the scene image in order to highlight salient regions, while the second branch processes the head crop of the target person to infer a general gaze direction. A fusion mechanism then combines information from both parts to produce the final prediction. 

The gaze following task is often framed as the prediction of a gaze heatmap where pixels with high intensity represent spatial areas with higher prediction confidence. Instead, the main variant of our Sharingan architecture directly regresses the 2D location of the gaze target.
Nevertheless by selecting appropriate decoders, we are also able to retain the traditional task formulation of predicting a heatmap.

\textbf{Multi-Person Gaze Following.} A major downside of the traditional formulation of gaze following is the need for multiple forward passes when predicting the gaze of different people in the same image. This is even more cumbersome when the gaze architecture requires multiple modalities in the input \cite{guan2020_pose, nan2021predicting, Fang_2021_CVPR_DAM, hu2022we, hu2022gaze}, leading to high computation costs for inference. This problem motivated the need for architectures that can natively handle the prediction of gaze for multiple people with a single forward pass. Jin \emph{et al.} \cite{jin2021multi} first proposed a simple convolution-based architecture to handle the multi-person setting where a scene backbone computes a fixed person-agnostic feature representation. This is then fused repetitively with head features computed from the different people using another head backbone before decoding each into its corresponding gaze heatmap. Aside from the architectural differences, one of the main limitations of this method is that the computation for each person is done independently from the others, which ignores the potential interactions between people. Recently, Tu \emph{et al.} \cite{tu2022end} proposed a transformer-based architecture to perform multi-person gaze target prediction. Their method only takes the image as input and simultaneously predicts both the head bounding box and corresponding gaze target for every person in the scene. Inspired by the DETR architecture \cite{carion2020end}, they formulate the gaze following task as a set prediction problem. Instead of reinventing the wheel, our method focuses solely on the gaze prediction part (\ie given that heads are easily and accurately obtainable using off-the-shelf detectors), and naturally adapts the transformer architecture to the task by introducing \emph{person tokens} alongside the standard image tokens found in a vision transformer \cite{dosovitskiy2020image_vit}. The \emph{person tokens} capture person-specific gaze and head location information and can be directly decoded into gaze predictions later in the architecture.

\textbf{Transformer Architecture.} Initially introduced for language translation \cite{vaswani2017attention}, the transformer architecture attracted a lot of interest in recent years. It has been widely adopted by different research communities (\eg text, vision, speech, multimodal) and successfully applied to a wide range of tasks \cite{dosovitskiy2020image_vit, carion2020end, liu2021swin, baevski2019effectiveness, devlin2018bert, radford2018improving, arnab2021vivit}. The transformer relies on an attention mechanism to dynamically attend to the relevant parts of the input. Thus, it effectively has a full receptive field from the early layers making it effective at capturing long-range dependencies. The ViT \cite{dosovitskiy2020image_vit} was the first attempt to adapt the transformer architecture to the vision domain, specifically to image classification. In order to build the set of tokens, the authors first split the input image into $16 \times 16$ non-overlapping patches which are then projected to an embedding space and equipped with positional information before going through the standard transformer blocks. 
The transformer encoder of our architecture is itself a ViT tha we 
simply extent to handle both scene and gaze related person tokens.

\textbf{Human-Human-Object Interaction.} Given its flexibility, Sharingan is meant to be a first step toward methods able to perform a multi-faceted analysis of social scenes by integrating different modalities (\eg image, depth, motion, semantics) and producing one or multiple desired outputs (\eg gaze, gestures, interactions, speaking status, etc.). Given that interaction is a fundamental component of social scenes, the Human-Human-Object Interaction (HHOI) detection task is close to this end goal, and prior works in this area can help inform architectural decisions for tasks related to social scene understanding. The goal of HHOI is to detect a source person and a target person or object being interacted with, as well as the nature of the interaction. Traditional methods to solve this task relied on multi-stream convolutional networks \cite{chao2018learning, gkioxari2018detecting, gao2018ican} to extract features from different people/objects produced by off-the-shelf detectors and the relational information between these entities. Later works found more success using graph neural network architectures \cite{qi2018learning, zhou2019relation, liang2021visual}. The task of HHOI naturally lends itself to a graph representation where the nodes represent the entities (\ie people, objects) and the edges represent the interactions between them. This formulation is also applicable to gaze prediction and has been attempted before \cite{hu2022gaze}. The major downsides of using graph neural networks however, is that the 2D spatial structure is lost in node representations, and off-the-shelf object detectors are often not able to detect all the various candidate objects in the scene which are valid gaze targets. Eventually, recent efforts in this area turned their attention to transformer-based architectures for HHOI detection \cite{zou2021end, kim2021hotr, zhou2022human, tu2022iwin} which were able to address some of those concerns.

%%%%%%%%% ARCHITECTURE
\section{Sharingan Architecture}
\label{sec:architecture}

% Architecture
\begin{figure*}
    \centering
    \includegraphics[width=0.95\textwidth]{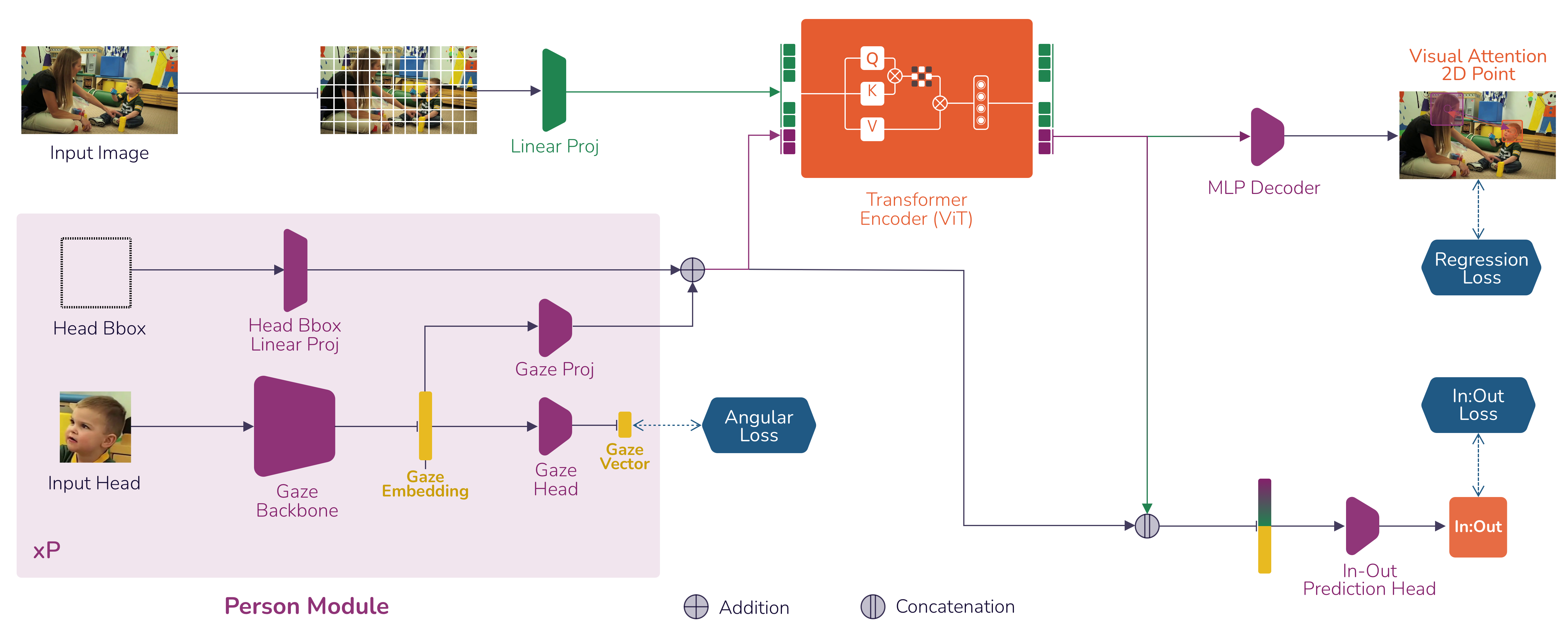}
    \caption{
      Overview of our proposed multi-person Sharingan architecture. Similar to ViT \cite{dosovitskiy2020image}, the input image is first split into non-overlapping patches which are then projected and equipped with 2d positional information to create image tokens (\textcolor[HTML]{208450}{green squares}). Next, for each person, the head box coordinates are projected to the dimension of the expected tokens, and the head crop is fed to a gaze backbone to produce a gaze embedding. This will be: 1. used to predict a normalized 2d gaze vector that is supervised using an angular loss, and 2. projected to the token dimension to produce a gaze token. The gaze token and head box embedding are summed to create a location-aware person gaze token (\textcolor[HTML]{84206B}{purple squares}). The image tokens together with the person tokens are fed to the transformer encoder, and the output tokens corresponding to input people are decoded using an MLP to regress the $(x, y)$ gaze point coordinates. Finally, the person gaze token is combined together with the corresponding output person token to predict the inside-vs-outside label.
    }
    \label{fig:architecture}
\end{figure*}

Our Sharingan architecture is illustrated in Figure \ref{fig:architecture}. 
The main idea is to use a transformer that let scene tokens and person-specific gaze tokens interact within an attention based architecture in order to regress for each individual the 2D gaze target location within the image. 
The main input are thus an image \inputimage 
$\in \mathbb{R}^{H \times W \times C}$ as well 
as the crops of the heads and faces that we assume 
have been detected. 
In the following, we introduce the different components of this architecture.

% 1. Primer on Transformers for Vision
%\subsection{Overview}

\subsection{Image tokens}

We follow a standard ViT architecture to produce inmage tokens.
The scene image \inputimage is first split into $P \times P$ non-overlapping patches which are flattened into a set \seqimgpatches $\in \mathbb{R}^{N \times (P^2 \cdot C)}$ of patch vectors where $N = \frac{H}{P} \cdot \frac{W}{P}$ is the number of image patches.
These are then fed to a learnable linear projection layer \imageproj to produce a set of $N$ image tokens \seqimgtokens $\in \mathbb{R}^{N \times D}$
where $D$ denotes the dimension of each token. 
We also add a non-learnable sine-cosine positional encoding in order to retain positional information resulting in the final token representation of the image $\imgtokens \in \mathbb{R}^{N \times D}$.

\subsection{Person gaze tokens}

Figure \ref{fig:architecture} depicts the gaze branch (person module)   
that is applied to each individual head crop to produce a gaze token. 
Its main purpose is to map the gaze information of a person into a token that lie in the same space (albeit with a different bias) than the image tokens, and which can interact with the scene tokens  \seqimgtokens to select the relevant content for regressing the gaze location.

\mypartitle{Single person case.}
Let \inputheadcrop $\in \mathbb{R}^{h \times w \times C}$ denote the head crop of a person and  \inputheadbbox $= (x_{\min}, y_{\min}, x_{\max}, y_{\max}) \in [0, 1]^{4}$ her head bounding box.
The mapping works as follows.
The head crop \inputheadcrop is fed to a gaze backbone \gazebackbone to produce a gaze embedding $\gazeemb \in \mathbb{R}^{d_{\text{emb}}}$. 
This embedding is used in two ways. 
First, it goes through a gaze prediction Multi-Layer-Perception (MLP) $\gazepred$ to predict a 2D gaze vector \gazevector: $\gazevector=\gazepred({\gazeemb})$. 
This part of the network will be used for defining a gaze loss. 

Secondly,  the gaze embedding is projected to the token dimension using a learnable linear projection \gazeproj, resulting in the gaze vector 
$\gazetoken = \gazeproj (\gazeemb) \in \mathbb{R}^{D}$. 
As we want to incorporate information about the person location (and size), we also project the head bounding box \inputheadbbox 
into the token dimension using a learnable linear projection \bboxproj: $\bboxemb = \bboxproj (\inputheadbbox) \in \mathbb{R}^{D}$. 
Finally, we sum the gaze  and head vectors to obtain the gaze token, \ie 
the location-aware representation of the 
person's gaze:
\begin{equation}
    \persontoken = \gazetoken + \bboxemb \in \mathbb{R}^{D}
\end{equation}

\mypartitle{Multi-person case.}
When \Npersons persons are detected, the architecture will produce a set of \Npersons gaze token, following exactly the same process described above for each person. 
Thus, if $\mathbf{h}_{\text{bbox}}^{i}$ and $\mathbf{h}_{\text{crop}}^{i}$ denote the bounding-box and head crop of person $i$, the above process will generate a person gaze token  $\persontoken_i$ for this person. 
With abuse of notation, we will also denote by $\persontoken$ the set 
of gaze tokens of all people in the scene, with  
$\persontoken = \persontoken_1\oplus \ldots \oplus\persontoken_{\Npersons}$, 
 where $\oplus$ denotes the concatenation operator.

%In order to achieve this, we consider $P$ people in the scene and their corresponding head bounding boxes  and  where $i \in \{1:P\}$. Each head crop $\mathbf{h}_{\text{crop}}^{i}$ is fed to the same gaze backbone \gazebackbone and produces a gaze embedding $\gazeemb_i$ which is then projected to $\gazetoken_i = \gazeproj(\gazeemb_i)$. Similarly, the corresponding head bounding box is projected to $\bboxemb_i = \bboxproj(\inputheadbbox^i)$, and the person token can be obtained $\persontoken_i = \gazetoken_i + \bboxemb_i$. This way, the input sequence to the transformer encoder becomes $\inputtokens = \imgtokens \oplus \persontoken_1 \oplus ... \oplus \persontoken_P \oplus \globaltoken \in \mathbb{R}^{(N+P+1) \times D}$. Now each output token $\outputtokens_i$ corresponding to person $i \in \{N+1:N+P\}$ can be decoded into $\gazepoint^i = \mlpdecoder (\outputtokens_i)$. 

\mypartitle{Token modality.}
Given the different nature of the gaze tokens compared to the image tokens, one may wish to encode modality specific information to distinguish between them. 
Rather than using an explicit scheme, in practice we expect this modality information to be captured by the bias terms of the different projector operators  \gazeproj and \imageproj.

% 1. Primer on Transformers for Vision
\subsection{Transformer Encoder}

The transformer encoder is a standard ViT \cite{dosovitskiy2020image}. 
It takes as input the concatenation of the image tokens \imgtokens, the gaze  token(s) \persontoken and a global token \globaltoken (\ie often referred to as the class token), according to $\inputtokens = \imgtokens \oplus \persontoken \oplus \globaltoken \in \mathbb{R}^{\Nalltokens \times D}$, 
where $\Nalltokens = \Nimagetokens + \Npersons + 1$.
The role of the global token is to aggregate and distribute information across the set of token.
The set of input tokens goes through a series of $L$ transformer blocks to obtain an output sequence of similar shape, denoted by $\outtokens = \token^{(L)} \in \mathbb{R}^{\Nalltokens \times D}$. 
Each transformer block comprises a multi-head self-attention followed by a feed-forward network, including a layer norm and a residual connection after each operation. We refer the interested reader to the original papers \cite{vaswani2017attention,dosovitskiy2020image} for more details.

% =============================================
\subsection{Decoder}
The goal of the decoder is to transform the output tokens $\outputtokens$ into a suitable prediction for the gaze following task.
There are several ways to do so, and we experimented with two variants:
\begin{compactitem}
\item Heatmap variant. It  follows the traditional task formulation of predicting a heatmap from all output tokens where the maximum indicates the predicted 2D gaze point. 
Its main drawback is to only support predicting the gaze of a single person for each forward pass.
\item 2D point regression. 
It casts the gaze following task as regressing the $(x, y)$ coordinates of the gaze point of one person from the output token of that person. The main benefits are to support multi-person prediction as well as accounting for person gaze interactions (\eg looking at each other) during learning and prediction.  
\end{compactitem}
More details about these two variants, including further discussions about the benefits and drawbacks of the methods are provided below.

% 1. Primer on Transformers for Vision
% \subsubsection{Heatmap prediction variant}
\mypartitle{Heatmap prediction variant.}
In this case (not shown in Figure~\ref{fig:architecture}), 
we assume that the gaze token of only one person is provided, 
and we generate the heatmap (\heatmap) by decoding the output tokens 
$\outputtokens_{\{1:N\}}$ corresponding to the $N$ input image tokens \imgtokens in \inputtokens. 
This is the transformer equivalent of decoding CNN feature maps produced from the scene image, fused with gaze information, 
which is how most previous works go about solving this task \cite{nips15_recasens, chong2020dvisualtargetattention, Fang_2021_CVPR_DAM, gupta2022modular, jin2021multi, lian2018believe, jin2022depth}. 
The rationale is that image tokens will be updated by the transformer through the attention mechanism to highlight candidate regions in the scene image where the target person might be looking. It also makes sense to decode a heatmap image using tokens that implicitly contain the 2D support structure of the original scene image. 

Since a heatmap pertains to a dense prediction task, 
we decided to use a Dense Prediction Transformer  (DPT) 
\cite{ranftl2021vision} as decoder. 
In brief, the DPT decoder reassembles tokens from different layers of the transformer encoder into image-like feature map representations at different resolutions where lower resolutions correspond to deeper layers of the encoder, and vice-versa.
These "feature maps" are then progressively combined and "upscaled" using convolution-based fusion modules until we obtain a full-resolution prediction. See the original paper of DPT \cite{ranftl2021vision} for more details.

One benefit of decoding from image-based tokens to predict the heatmap is 
that image tokens learn person-specific patterns through their interaction with the person token, and that the heatmap can highlight different modes in the posterior distribution when more than one gaze target is probable.

% 1. Primer on Transformers for Vision
\mypartitle{2D point regression variant}
In this case, the aim is to predict the gaze target point of person $i$ by decoding the output token $\outtokens_{N+i}$ of that person.  
As this  token originates from the head crop image pooled into 
a 1D representation, it may dilute the 2D spatial structure, even though it can interact with all the different image tokens.
Hence decoding the person token as a gaze heatmap might be challenging.  Instead, we prefer to directly regress the 2D gaze location  by using an MLP decoder \mlpdecoder, 
\ie $\gazepoint^i = \mlpdecoder(\outtokens_{N+i})$.

The two main advantages of this approach are 
(i) to allow \emph{multi-person prediction} in one forward pass, and 
(ii) modeling \emph{multi-person interaction} both at training and inference time since person tokens $\persontoken_i$ can interact with one another.
This is particularly important in social scenes where there is often a strong inter-dependency between head and gaze information 
of interacting people (\eg shared attention, looking at each other).
A disadvantage is that image tokens at different layers of the transformer may have to capture all scene salient items that may be relevant to any visible person. In other words, the inferred image features and tokens cannot be specific to a single person.

\subsection{In-Out prediction}

The In-Out prediction head \inoutmlp consists of an MLP with 7 layers. 
It takes as input the concatenated person output token $\outtokens_i$ and gaze token $\persontoken_i$ to predict a binary in-vs-out of frame gaze label for person $i$.
\begin{equation}
    \inoutlabel_i = \inoutmlp([\outtokens_i, \persontoken_i])
\end{equation}

A value of $1$ indicates that the person is looking at an item inside the scene image, whereas a value of $0$ indicates that the person is looking outside the scene image.

\subsection{Loss and implementation details}

We train our model using a combination of three losses that define our global loss $\loss$.

\mypartitle{Regression Loss ($\loss_{reg}$).} 
It has two variants corresponding to the Sharingan heatmap and 2D point regression models. For the heatmap model, we compute the pixel-wise MSE loss between the GT heatmap and the predicted heatmap:
$\loss_{hm} = \sum_{x,y}^{W_{hm},H_{hm}} || \heatmap^{\text{gt}}_{x,y} - \heatmap^{\text{pred}}_{x,y} ||_2^2$. 

For the 2D point regression model, we compute the distance-wise 
MSE between the predicted and GT gaze point locations :
$\loss_{pt} = || \gazepoint^{\text{gt}} - \gazepoint^{\text{pred}} ||_2^2$.

\mypartitle{Angular Loss ($\loss_{ang}$).}
The angular loss drives to a large extend the gaze backbone. 
Its maximizes the cosine of the angle between the predicted and ground truth gaze vectors according to:
$\loss_{ang} = 1 - <\gazevector^{gt}, \gazevector^{pred}>$ where $<a,b>$ denotes the inner product between $a$ and $b$.

\mypartitle{In-Out Loss ($\loss_{io}$).} 
The in-out loss is the standard binary cross-entropy loss between the predicted and ground truth in vs out of frame gaze labels.

\mypartitle{Global loss.}
The final loss is a linear combination of the three losses:
\begin{equation}
    \loss = \losscoeff_{reg} \loss_{reg} + \losscoeff_{ang} \loss_{ang} + \losscoeff_{io} \loss_{io}
\end{equation}

%%%%%%%%% EXPERIMENTS
\section{Experiments}
\label{sec:experiments}
% Experiments

\subsection{Datasets}

We test our models on two public benchmarks.

\mypartitle{GazeFollow.}
GazeFollow~\cite{recasens2017following} is an image based dataset consisting of images curated from popular image benchmarks such as COCO~\cite{lin2014coco}. The dataset is annotated with head bounding boxes, 2D gaze points and in vs out of frame gaze labels (in vs out labels provided by~\cite{chong2018connecting}). 
Overall, it has annotations for 130K people in 122K images. 
The test set comprises 4782 gaze instances (all inside the image) 
with 2D gaze points marked by 10 annotators. 

\mypartitle{VideoAttentionTarget.}
VideoAttentionTarget~\cite{chong2020dvisualtargetattention} is a video based dataset consisting of 1331 clips from 50 TV shows. The dataset is also annotated with the head bounding boxes, 2D gaze points and in vs out of frame gaze labels. Overall, it has annotations for 164K people in 71K frames. 

\subsection{Experimental protocol}

\mypartitle{Implementation Details.} 
Sharingan processes the input scene image and head crop at a resolution of $W \times H = 224 \times 224$, while the output heatmap (when using a heatmap) 
has a resolution of $W_{hm} \times H_{hm} = 64 \times 64$. 
The gaze backbone \gazebackbone is a ResNet-18~\cite{he2016deep_resnet} 
pre-trained on Gaze360~\cite{kellnhofer2019gaze360}. 
The transformer encoder is a ViT~\cite{dosovitskiy2020image_vit} 
Base model initialized with weights from Bachmann et al.~\cite{bachmann2022multimae}.

\mypartitle{Training.} 
The models are trained for 30 epochs on GazeFollow. For VideoAttentionTarget, we take the trained GazeFollow model and fine-tune it for another 20 epochs. We use the AdamW optimizer~\cite{loshchilov2018decoupled_adamw} with a learning rate of $3e-5$ cosine annealing with warm restarts \cite{loshchilov2016sgdr} as a learning rate schedule. We also make use of Stochastic Weight Averaging~\cite{izmailov2018averaging_swa} to stabilize training. The loss coefficients are $\lambda_{reg} = 1000$ for the heatmap, $\lambda_{reg} = 100$ for the 2D point regression, and $\lambda_{ang} = 3$ for both.

\mypartitle{Validation.} Since GazeFollow~\cite{recasens2017following} and VideoAttentionTarget~\cite{chong2020dvisualtargetattention} do not propose any validation split, we split a portion of the training set and use it for validation. Our GazeFollow validation split consists of 4499 instances, while our VideoAttentionTarget validation split consists of 6726 instances from 3 shows. The best model as per the validation set is used for testing.

\subsection{Tested models}

\mypartitle{Models.} We train and evaluate three models:
\begin{itemize}
    \item Sharingan heatmap variant (Heatmap): This model predicts the gaze target for a single person in the form of a heatmap.
    \item Single person 2D point variant (2D point, \Npersons=1): This version of the 2D point model is trained and evaluated with $\Npersons=1$ person token. There is no person-person interaction present in this model.
    \item Multi-person 2D point variant (2D point, \Npersons=6): This version of the 2D point model is trained and evaluated with a $\Npersons=6$ person tokens. If there are less than 6 people in an image, we provide black images as extra heads.
\end{itemize}

\mypartitle{Ablation.} We evaluate the performance of each model when tested with different numbers of people as input (different from the \Npersons they were trained with).

\begin{table*}[t]
    \centering
    \begin{tabular}{l c c c c c c c c}
    \hline
      &  &  & \multicolumn{3}{c}{\textbf{GazeFollow}} & \multicolumn{3}{c}{\textbf{VideoAttentionTarget}} \\
    \textbf{Model} & \textbf{Type} & \textbf{Modalities} & \hspace{-2mm}\textbf{Avg. Dist}$\downarrow$ & \hspace{-2mm}\textbf{Min. Dist}$\downarrow$ & \hspace{-2mm}\textbf{AUC}$\uparrow$ & \hspace{-2mm}\textbf{Dist}$\downarrow$ & \hspace{-2mm}\textbf{AUC}$\uparrow$ & \hspace{-2mm}\textbf{AP}$\uparrow$ \\
    \hline 
    \hline
    Recasens~\cite{nips15_recasens} & single & image & 0.190 & 0.113 & 0.878 & - & - & - \\
    Lian~\cite{lian2018believe} & single & image & 0.145 & 0.081 & 0.906 & - & - & - \\
    Chong~\cite{chong2020dvisualtargetattention} & single & image & 0.137 & 0.077 & 0.921 & 0.147 & 0.854 & 0.848 \\
    Fang~\cite{Fang_2021_CVPR_DAM} & single & image+depth+eyes & 0.124 & 0.067 & 0.922 & \textbf{\textcolor{blue}{0.108}} & 0.905 & 0.896 \\
    Fang~\cite{Fang_2021_CVPR_DAM} & single & image+depth & - & - & - & 0.124 & 0.878 & 0.872 \\
    Jin~\cite{jin2022depth} & single & image+depth & 0.118 & 0.063 & 0.920 & 0.109 & 0.898 & \textbf{\textcolor{blue}{0.897}} \\
    Jin~\cite{jin2022depth} & single & image & 0.137 & 0.077 & 0.909 & - & - & - \\
    Gupta~\cite{gupta2022modular} & single & image+depth+pose & 0.114 & 0.056 & \textbf{\textcolor{blue}{0.943}} & 0.110 & 0.913 & 0.879 \\
    Gupta~\cite{gupta2022modular} & single & image & 0.134 & 0.071 & 0.933 & 0.122 & \textbf{\textcolor{blue}{0.918}} & 0.864 \\
    Hu~\cite{hu2022gaze} & single & image+depth+objects & 0.128 & 0.069 & 0.923 & 0.118 & 0.880 & 0.881 \\
    Jin~\cite{jin2021multi} & multi & image & 0.126 & 0.076 & 0.919 & 0.134 & 0.881 & \textbf{\textcolor{red}{0.880}} \\
    Tu~\cite{tu2022end} & multi & image & 0.133 & 0.069 & 0.917 & 0.137 & 0.893 & 0.821 \\
    \hline
    \hline
    \rowcolor{LightCyan}
    Ours (Heatmap) & single & image & 0.108  & \textbf{\textcolor{blue}{0.054}} & 0.938 & 0.113 & 0.831 & 0.823 \\
    \rowcolor{LightCyan}
    Ours (2D Point, $\Npersons=1$) & single & image & \textbf{\textcolor{blue}{0.104}} & 0.064 & - & 0.112 & - & 0.857 \\
    \rowcolor{LightCyan}
    Ours (2D Point, $\Npersons=6$) & multi & image & \textbf{\textcolor{red}{0.106}} & \textbf{\textcolor{red}{0.066}} & - & \textbf{\textcolor{red}{0.118}} & - & 0.854 \\
    \hline
    \hline
    Human & - & - & 0.096 & 0.040 & 0.924 & 0.051 & 0.921 & 0.925 \\
    \hline \\
    \end{tabular}
    
    \caption{Results of our Sharingan variants on the GazeFollow and VideoAttentionTarget datasets. The best scores for the single-person models are given in \textcolor{blue}{blue}, and the best scores for the multi-person models are given in \textcolor{red}{red}. In the sharingan 2D point variant, $n$ refers to the total number of people used for training and evaluation. Also, the results reported on VideoAttentionTarget represent the corresponding GazeFollow pre-trained models where we only fine-tune the in-vs-out classifier.} 
    \label{tab:gazefollow-results}
\end{table*}

\subsection{Results}

Our quantitative results on the GazeFollow and VideoAttentionTarget datasets compared to previous works are summarized in Table \ref{tab:gazefollow-results}.

\mypartitle{GazeFollow results.}
The Heatmap variant of the Sharingan architecture achieves state-of-the-art results on GazeFollow across both Avg. and Min. Distance metrics by a healthy margin, even when compared to methods using multiple modalities as input. It falls slightly behind \cite{gupta2022modular} 
which exploits 3 input modalities in terms of AUC, but it is worth noting that this metric is relatively difficult to interpret, 
and the values obtained are already better than human performance, unlike distance-based metrics.

The multi-person 2D point variant of our architecture also achieves SOTA results across both applicable metrics compared to other multi-person models. The Avg. Distance in particular even shows an improvement in contrast to our Heatmap variant, but the Min. Distance is worse. 
This is a pattern that we noticed consistently with all our models trained by regressing the $(x, y)$ gaze coordinates directly, where the Avg. Distance improves compared to the Heatmap models, 
but Min. Distance slightly degrades. 
%remains relatively high. 
%
Indeed, since these model can only predict one value, 
we believe that these models converge to some form of 
the expectation of the posterior probability. 
When this distribution is multi-modal (\ie there is more than one probable gaze target), the expectation can become unlikely under that posterior distribution. 
This might explain why for these 2D point regression variants the Avg. Distance is slightly lower given that it literally represents the distance to the ground-truth average point. 
In contrast, Heatmap models do not suffer from this issue because the predicted intensity map is able to capture the different modes of the distribution, and taking the $\arg\max$ is essentially equivalent to selecting the 2D point maximizing the posterior.

Finally, the single-person variant of our architecture exhibits the best Avg. Dist. performance, while, as explained above, it also suffers from lower Min. Dist. Performance.
One explanation why this model performs better 
is that in the transformer, the processing of
the image token may specialize specifically 
to identify the salient items relevant 
to the person we are  estimating the gaze from. 

\mypartitle{VideoAttentionTarget (VAT) results.}
For this dataset, we report results of the model trained on GazeFollow, with only the in-vs-out classifier being trained on the VAT data. 
Indeed, the different attempts at fine-tuning the whole model on VAT, as is commonly done, did not improve the results. 
This might be due to the lack of diversity of this dataset, 
and hence large models like transformers may overfit the data 
or may not benefit from it. 

Nevertheless, our Heatmap models demonstrates good cross-dataset performance, having the best results when using only the image as input modality. 
Furthermore, our multi-person model beats other models of the same nature by a good margin as well, demonstrating also its generalization capacity. 

\begin{figure}
    \centering
    \includegraphics[width=0.95\linewidth]{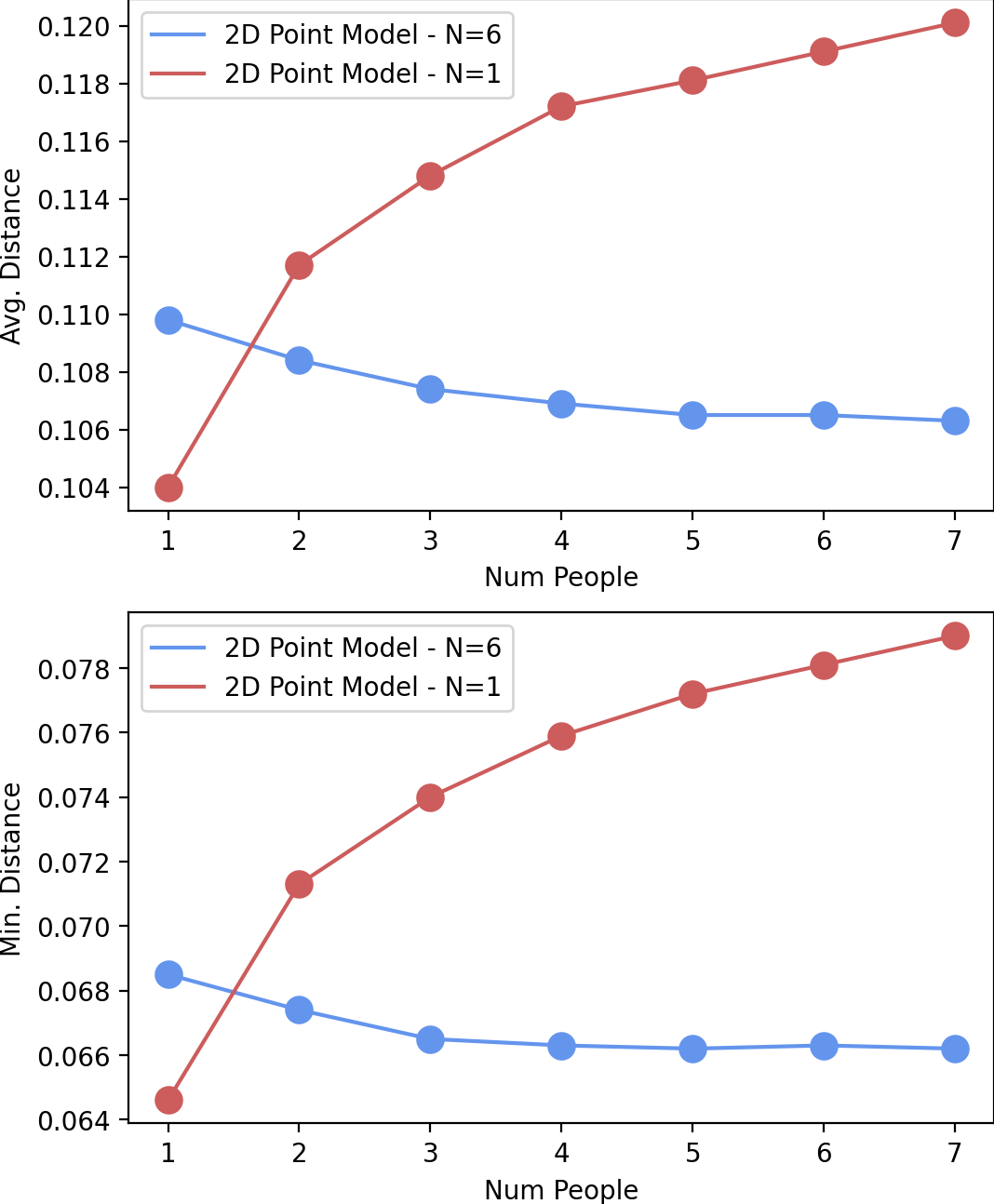}
    \caption{
      Performance comparison between a Sharingan (2D Point) model trained using $\Npersons=1$ and $\Npersons=6$ across different values of $\Npersons$ for evaluation. Results are reported on the test set of GazeFollow. 
    }
    \label{fig:ablation}
\end{figure}

\begin{figure*}
    \centering
    \includegraphics[width=0.95\linewidth]{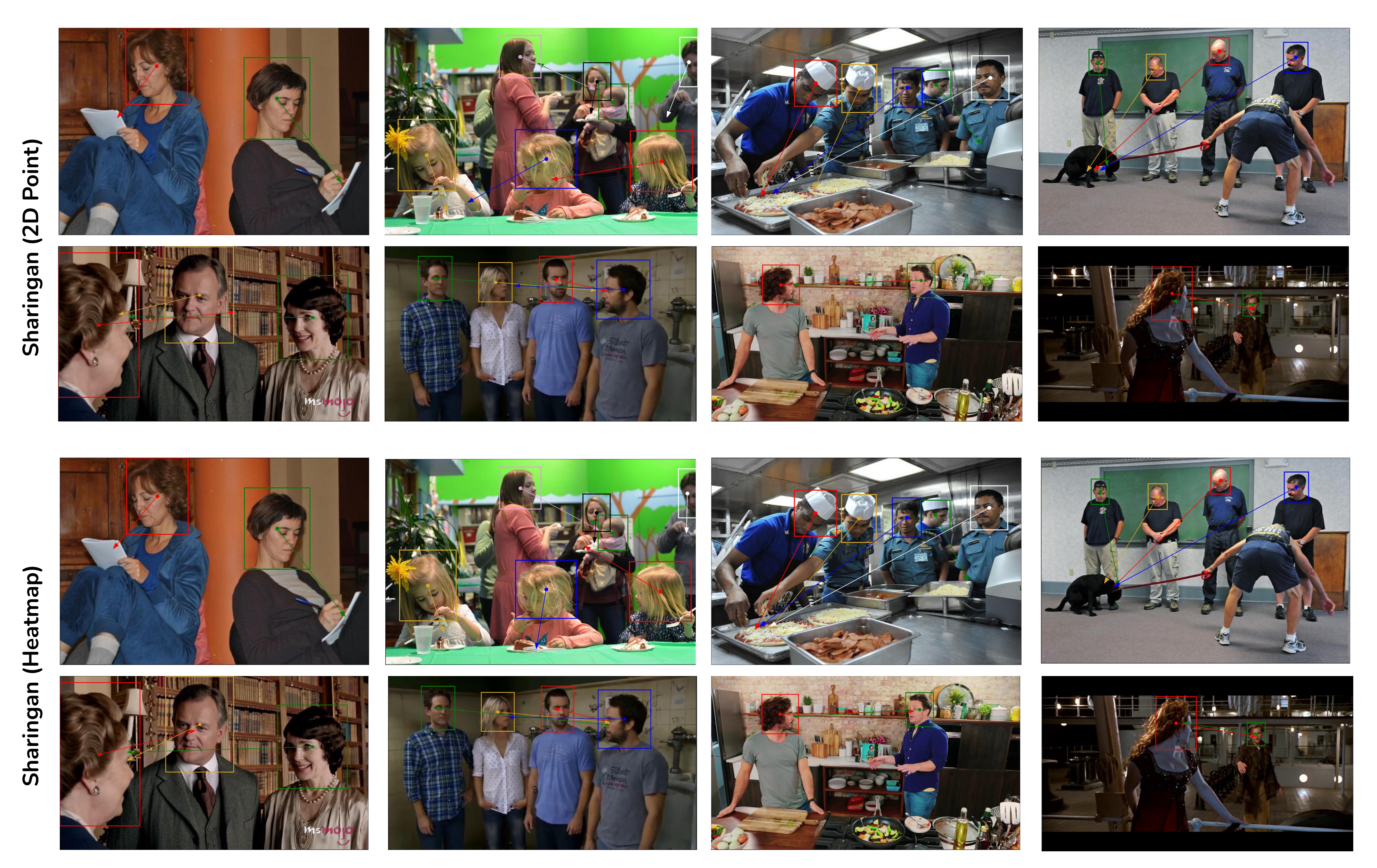}
    \caption{
      A random sample of qualitative results comparing the Sharingan (heatmap) and Sharingan (2D Point, $\Npersons=6$) variants. The first row is selected from the test set of GazeFollow while the second row represents the test set of VideoAttentionTarget. Both of the models showcased here are only trained on GazeFollow. We use an off-the-shelf head detector to extract people to feed into the model.
    }
    \label{fig:qualitative}
\end{figure*}

\mypartitle{Ablation.} We plot the results of testing with different numbers of people as input for the 2D point variants in Figure~\ref{fig:ablation}. We see that the single person 2D point variant has the best performance for a single person as input, and degrades in performance as we provide more people as input. This is reasonable as the model has not learned the interactions between multiple people. Further, as discussed previously, in this model the image tokens may specialize to the salient items relevant to the input person. As such, with more than one person as input this specialization is hindered. 

For the multi person 2D point variant, we see increasing performance with an increase in the number of people as input (up to \Npersons=5). This is because the model can leverage more person-person interactions in the scene. Beyond \Npersons=5 there may not be additional cues that the model can benefit from hence we do not see further improvement. The largely stable results for \Npersons lower and higher than what the model has been trained for highlight the value of our model for evaluation under different settings.

\mypartitle{Qualitative Results.} We show the qualitative results from our models in Figure~\ref{fig:qualitative}. The models were trained on GazeFollow, and tested on images from GazeFollow (first row) and VideoAttentionTarget (second row). We note generally good performance for both the Heatmap and multi-person 2D point model. Importantly, the multi-person model provides comparable performance to the Heatmap model at a fraction of the inference cost.

%%%%%%%%% CONCLUSION
\section{Conclusion}
\label{sec:conclusion}

% Conclusion
In this paper we proposed a new transformer based architecture for gaze target prediction: Sharingan. The first variant processes a single person and predicts the gaze target as a standard heatmap, achieving the new state of the art on GazeFollow. The second is a novel variant that predicts the gaze target as a 2D point. An important feature of this model is its support multiple people as input. Our experiments show that this model benefits from training and evaluating with multiple people, effectively learning person-person interactions in the scene. At the same time, it achieves the new state of the art for multi-person gaze target prediction on GazeFollow and VideoAttentionTarget. Its performance is also comparable to the state of the art single person models while performing inference at a fraction of their cost. In the future, we plan to  extend this model with multimodal cues for effective social scene understanding.

{\small
\bibliographystyle{ieee_fullname}
\bibliography{sharingan_final}
}

\end{document}